\documentclass[10pt,conference]{IEEEtran}

\usepackage{cite}
\usepackage{balance}

\ifCLASSINFOpdf
   \usepackage[pdftex]{graphicx}
   \graphicspath{{figs/}}
   \DeclareGraphicsExtensions{.pdf,.jpeg,.png}
\else
   \usepackage[dvips]{graphicx}
   \graphicspath{{../figs/}}
   \DeclareGraphicsExtensions{.eps}
\fi

\usepackage[cmex10]{amsmath}
\usepackage{amsthm}

\usepackage{algorithmic}
\usepackage{array}
\ifCLASSOPTIONcompsoc
  \usepackage[caption=false,font=normalsize,labelfont=sf,textfont=sf]{subfig}
\else
  \usepackage[caption=false,font=footnotesize]{subfig}
\fi

\usepackage{url}
\usepackage{multirow,ctable}

\begin{document}
%
\title{Activity Recognition based on a Magnitude-Orientation Stream Network}
\newif\iffinal
\finaltrue
\newcommand{\jemsid}{92}
\newcommand{\todo}[1] {{\color{red}[\textbf{TODO: }#1]}}
\newcommand{\alterar}[1] {{\color{red}[\textbf{}#1]}}
\newcommand{\verify}{{\color{red} \textbf{[*verify*] }}}
\newcommand{\verified}{{\color{blue} \textbf{[*verified - ok*] }}}
\newcommand{\changed}[1] {{\color{blue}[\textbf{changed: }#1]}}


\iffinal


\author{\IEEEauthorblockN{Carlos Caetano,
		Victor H. C. de Melo,
		Jefersson A. dos Santos, 
		William Robson Schwartz}
	\IEEEauthorblockA{Smart Surveillance Interest Group, Department of Computer Science\\
		Universidade Federal de Minas Gerais, Belo Horizonte, Brazil}
	\{carlos.caetano,victorhcmelo,jefersson,william\}@dcc.ufmg.br
}

\else
  \author{Sibgrapi paper ID: \jemsid \\ }
\fi

\maketitle

\begin{abstract}
	The temporal component of videos provides an important clue for activity recognition, as a number of activities can be reliably recognized based on the motion information. In view of that, this work proposes a novel temporal stream for two-stream convolutional networks based on images computed from the optical flow magnitude and orientation, named Magnitude-Orientation Stream (MOS), to learn the motion in a better and richer manner. Our method applies simple non-linear transformations on the vertical and horizontal components of the optical flow to generate input images for the temporal stream. 
	Experimental results, carried on two well-known datasets (HMDB51 and UCF101), demonstrate that using our proposed temporal stream as input to existing neural network architectures can improve their performance for activity recognition.
	Results demonstrate that our temporal stream provides complementary information able to improve the classical two-stream methods, indicating the suitability of our approach to be used as a temporal video representation.
\end{abstract}

\IEEEpeerreviewmaketitle

\section{Introduction}\label{introduction}

Human activity recognition has been used in many real-world applications. In environments that require a higher level of security, surveillance systems can be used to detect and prevent abnormal or suspicious activities such as robberies and kidnappings. In addition, human activity recognition can be employed in systems for video retrieval, so that a user is able to search for videos containing specific activities. Another type of application is in health care, such as activities of daily living  monitoring systems.

Surveillance applications have traditionally relied on network cameras monitored by human operators that must be aware of the activities carried out by people who are in the camera field of view. With the recent growth in the number of cameras to be analyzed, the efficiency and accuracy of human operators has reached the limit~\cite{Keval:2006}. Therefore, security agencies have attempted computer vision-based solutions to replace or assist the human operator. Automatic recognition of suspicious activities is a problem that has attracted the attention of researchers in the area~\cite{Danafar:2007, Reddy:2011, Wiliem:2012, WangAndXu:2016}.

A significant portion of the progress on activity recognition task has been achieved with the design of discriminative feature descriptors exploring temporal information. Such information is based on motion analysis and is very important to represent the video in a more discriminative space, allowing the improvement of activity recognition.

Over the last decade, most of the works focused on designing handcrafted local feature descriptors~\cite{Scovanner:2007, Laptev:2008, Klaser:2008, Wang:2011} or on encoding schemes using mid-level representations, such as Bag-of-Words (BoW)~\cite{Sivic:2003} or Fisher vector (FV)~\cite{Sanchez:2013}, followed by Support Vector Machines (SVM) classifier. Nowadays, large efforts have been directed to the employment of deep Convolutional Neural Networks (CNNs). 
These architectures learn hierarchical layers of representations to perform pattern recognition and have demonstrated impressive results on many pattern recognition tasks (e.g., image classification~\cite{Krizhevsky:2012} and face recognition~\cite{Schroff:2015}). Although the excellent improvements achieved in such tasks, activity recognition lacks on performance when using CNNs. Many works~\cite{Feichtenhofer:2016, Park:2016, Diba:2016} point that the potential reason behind such gap falls in two cases: (i) current datasets do not have enough videos for training and are too much noisy; and (ii) current CNN architectures are still not able to handle temporal information (or to take full advantage of it), consequently letting spatial (appearance) information prevail. 

A major breakthrough spatiotemporal information representation was achieved by Simonyan and Zisserman~\cite{Simonyan:2014}, who directly incorporated motion information by using optical flow instead of learning it from scratch, showing significant improvement over other approaches. Known as two-stream network, their architecture is composed of two stream of data: (i) spatial network, which takes as input the raw RGB pixels; and (ii) temporal network, which takes as input dense optical flow displacement fields (vertical and horizontal components) computed across the frames. The final predictions are computed as the average of the output scores from the two streams.

To further improve the representation of spatiotemporal information, this work introduces a new temporal stream for the two-stream networks to perform activity recognition, named \emph{Magnitude-Orientation Stream} (MOS). The method is based on non-linear transformations on the optical flow components to generate input images for the temporal stream. Our hypothesis is based on the assumption that the motion information on a video sequence can be described by the spatial relationship contained on the local neighborhood of magnitude and orientation extracted from the optical flow. More specifically, we assume that the motion information is adequately specified by fields of magnitude and orientation. In view of that, our method captures not only the displacement, by using orientation, but also magnitude providing information regarding the velocity of the movement. 

In the literature, magnitude and orientation information are often used to describe motion information  in various local  handcrafted-based features, such as Motion Boundary Histogram (MBH)~\cite{Dalal:2006}, Histogram of Oriented Flow (HOF)~\cite{Laptev:2008}, Histograms of Optical Flow Orientation and Magnitude (HOFM)~\cite{Colque:2015} and Optical Flow Co-occurrence Matrices (OFCM)~\cite{Caetano:2016}. However, none of the aforementioned methods used such information on an end to end learning scheme with a CNN, as the proposed approach does.


According to the experimental results, our proposed temporal stream used as input to existing neural network architectures is able to recognize activities accurately on two well-know datasets (UCF101~\cite{Soomro:2012} and HMDB51~\cite{Kuehne:2011}) outperforming the results achieved by the original two-stream network as well as other deep networks available in the literature.


\section{Related Works}\label{related}

\begin{figure*}[!t]
	\centering
	\begin{tabular}{>{\centering\arraybackslash}m{6cm} >{\centering\arraybackslash}m{6cm} >{\centering\arraybackslash}m{6cm}}
		\includegraphics[width=0.35\textwidth]{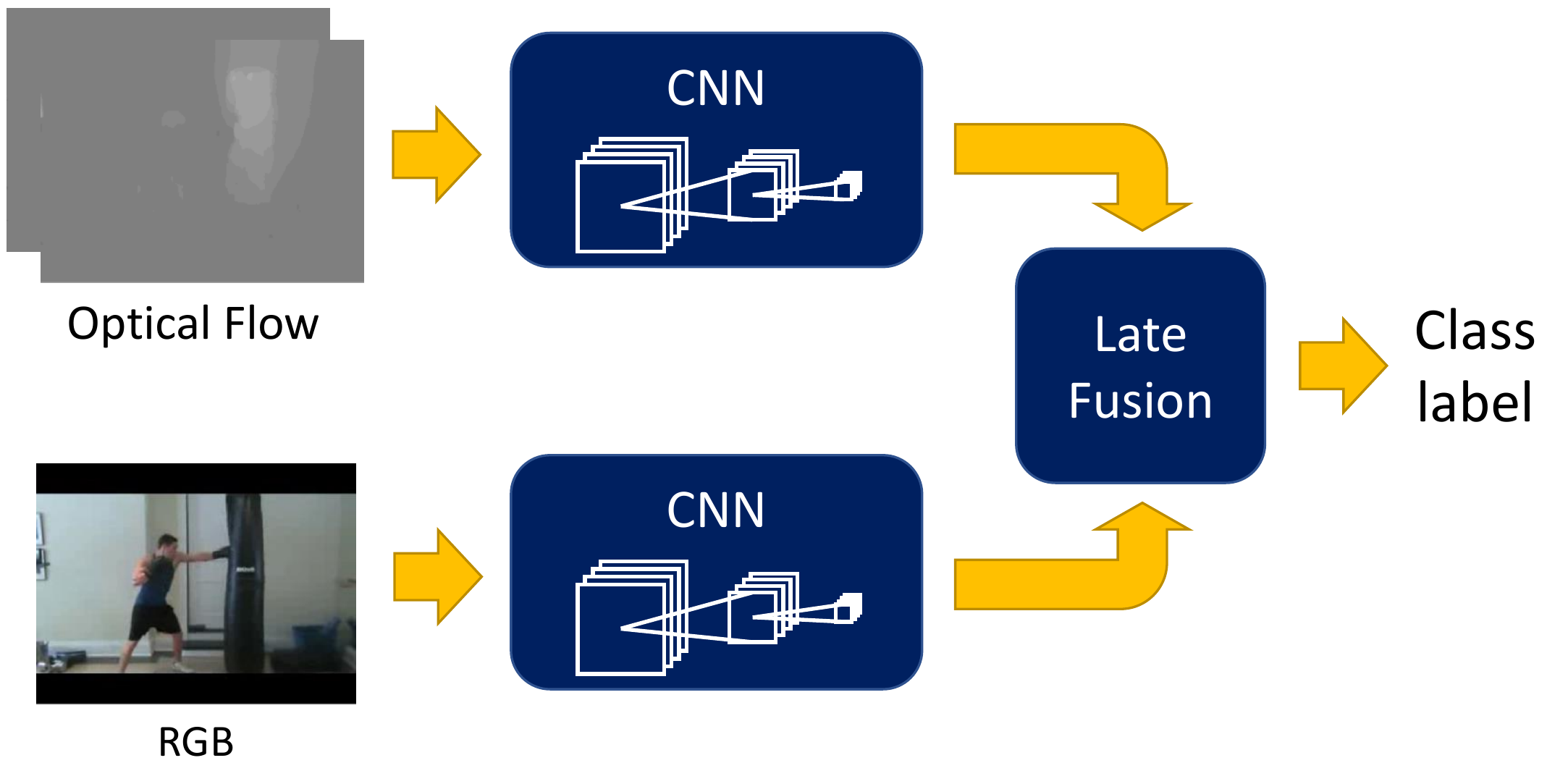} & \includegraphics[width=0.27\textwidth]{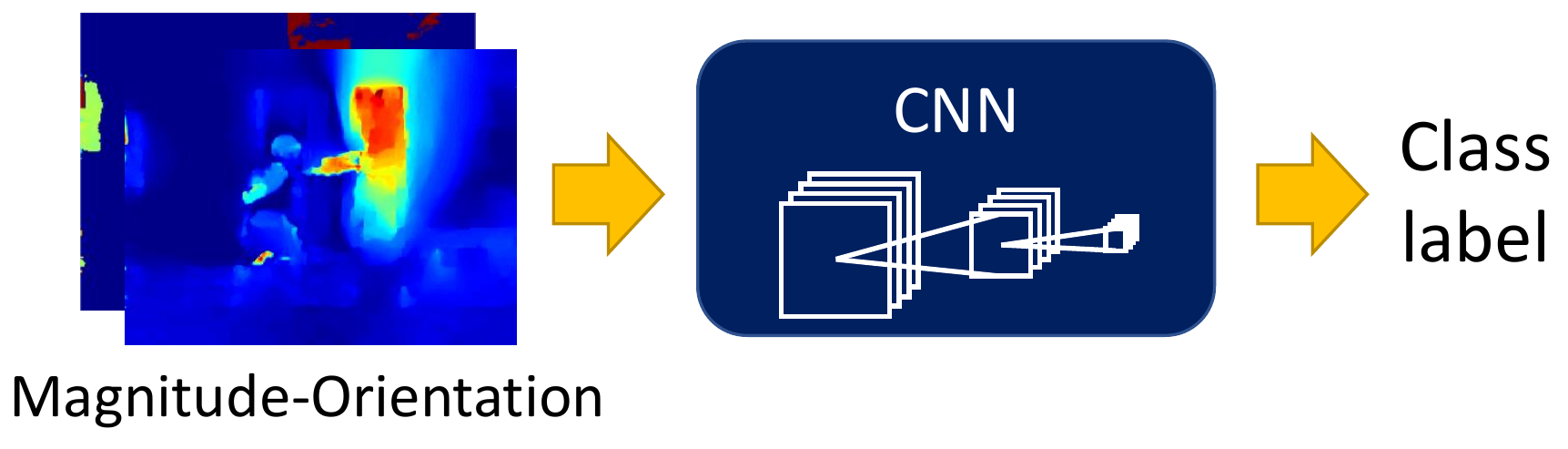} &	\includegraphics[width=0.35\textwidth]{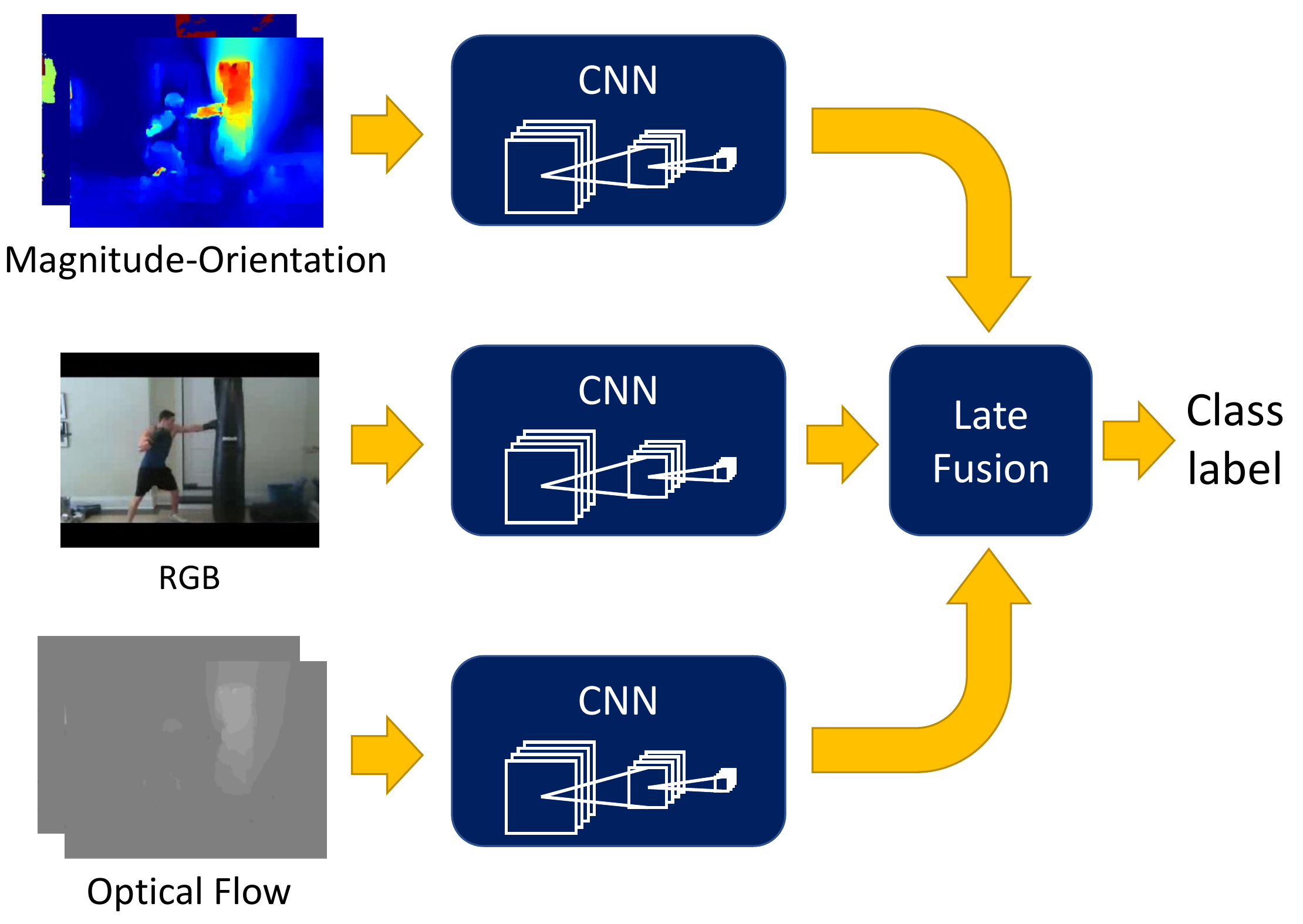}\\
		\footnotesize{(a) Very Deep Two-Stream~\cite{Wang:2015}.} & \footnotesize{(b) Our Magnitude-Orientation Stream} & \footnotesize{(c) Fusion of our approach and}\\
		& \footnotesize{(MOS) approach.} & \footnotesize{Very Deep Two-Stream~\cite{Wang:2015}.}\\
	\end{tabular}
	\caption{Architectures considered in this work for extracting spatiotemporal information.}
	\label{img:pipeline}
\end{figure*}


In this section, we present a literature review of works that are close to the idea proposed in our approach. These methods can be categorized by: (i) temporal information extracted from videos through the use of handcrafted local feature descriptors (Section~\ref{handcrafted}), and (ii) recent works that employ neural networks to learn temporal information (Section~\ref{deeplearning}).

\subsection{Methods based on Handcrafted Feature Descriptors}\label{handcrafted}


To characterize motion and appearance of local features, Laptev et al.~\cite{Laptev:2008} computed histogram descriptors of space-time volumes in the neighborhood of detected points. Each volume is subdivided into a grid of cuboids and, for each cuboid, they compute Histogram of Oriented Gradients (HOG)~\cite{Dalal:2005} and Histogram of Optical Flow (HOF). The HOG descriptor is computed by dividing the cuboid into regions and accumulating a histogram binned by gradient directions over the pixels, while HOF is binned according to the flow orientations and weighted by according its magnitude. Then, normalized histograms are concatenated and named HOG-HOF.

Dalal et al.~\cite{Dalal:2006} introduced the Motion Boundary Histogram (MBH). First applied to human detection, the motion boundary coding scheme captures the local orientations of motion edges based on HOG feature descriptors~\cite{Dalal:2005}. Treating the horizontal and vertical components of the optical flow as independent ``images'', the authors take their local gradients separately, find the corresponding magnitudes and orientations and use these as weighted votes to the local orientation histograms. Later on, the MBH was used on several works to describe motion information for activity recognition~\cite{Laptev:2008, Wang:2011, Wang:2013}.

The HOG feature descriptor was extended by Kl\"{a}ser et al.~\cite{Klaser:2008}, named as HOG3D. It is based on histograms of 3D gradient orientations computed using an integral video representation. The gradients are binned into regular polyhedrons in a multi-scale fashion in space and time. Therefore, HOG3D combines appearance and motion information.

Aiming at encoding both local static appearance and motion information, as in the HOG3D, but avoiding high dimensionality and a relatively expensive quantization cost, Shi et al.~\cite{Shi:2015} proposed the Gradient Boundary Histograms (GBH). Instead of using image gradients, the authors use time-derivatives of image gradients to emphasize moving edge boundaries. For each frame, they compute image gradients and apply temporal filtering over two consecutive gradient images. Then, they compute the magnitude and orientation for each pixel which are used to build a histogram of orientation as in HOG.

Colque et al.~\cite{Colque:2015} developed a feature called Histograms of Optical Flow Orientation and Magnitude (HOFM). Different from HOF that only encodes orientation information, HOFM captures the orientation and the magnitude of flow vectors providing information regarding the velocity of the moving objects. They build a 3D matrix based on the orientation and magnitude information provided by the optical flow field, where each line corresponds to a given orientation range and each column to the magnitude ranges. The authors then extended it to capture information regarding  appearance and density of regions by encoding the entropy of the orientation flow~\cite{Colque:2017}.

Aiming at capturing richer information from the optical flow, Caetano et al.~\cite{Caetano:2016} proposed the Optical Flow Co-occurrence Matrices (OFCM). The descriptor is based on the extraction of a set of statistical measures from co-occurrence matrices computed using the magnitude and orientation from optical flow information. Their hypothesis for designing the OFCM is based on the assumption that the motion information on a video sequence can be described by the spatial relationship contained on local neighborhoods of the flow field.

A major breakthrough on local feature-based approaches was achieved by Wang et al.~\cite{Wang:2011} which proposed an method to describe videos by dense trajectories. Trajectory shapes encode local motion information by tracking spatial interest points over time. To generate the trajectories, they sample interest points in space and time, and track them based on displacement information using an efficient dense optical flow algorithm. The HOG, HOF and MBH feature descriptors are used to describe the trajectories which are then encoded by Bag-of-Words (BoW) mid-level representation. Afterwards, the authors improved it to the Improved Dense Trajectories (IDT)~\cite{Wang:2013} using the homography between consecutive frames to estimate the camera motion and Fisher vector encoding. 


Although there are many approaches based on local feature descriptors, these works often require over engineering (e.g., feature extraction, mid-level representation and classifier training). On contrary, CNNs are a class of deep learning models that replace all engineering with a single neural network trained end to end from pixel values to classifier outputs~\cite{Karpathy:2014}.

\subsection{Methods based on Neural Network Approaches}\label{deeplearning}

Convolutional Neural Networks (CNNs) have achieved impressive state-of-the-art results on image classification~\cite{Krizhevsky:2012}. Therefore, many works have tried to apply CNNs to learn spatiotemporal information for activity recognition task. A natural choice, the 3D convolutional network was presented by Ji et al.~\cite{Ji:2013}, where they tried to learn both appearance and motion features with 3D convolution operations. Their method works by stacking consecutive segments of human subjects in videos and by applying 3D convolutions over such volume aiming that the first layer learns spatiotemporal features. Tran et al.~\cite{Tran:2015} also explored 3D CNNs. However, in contrast with Ji et al.~\cite{Ji:2013}, their method takes full video frames as inputs and does not rely on any preprocessing.

Karpathy et al.~\cite{Karpathy:2014} also used CNN aiming to learn motion features. The authors investigated different temporal information fusion schemes, learning local motion direction/speed with global information. Although significant gains in accuracy compared to the works based on handcrafted features, only little improvement was achieved when compared to single-frame CNN models, showing that the current CNN architectures are unable to efficiently learn motion features.

A major breakthrough was achieved by Simonyan and Zisserman~\cite{Simonyan:2014}. Instead of trying to learn motion information as Karpathy et al.~\cite{Karpathy:2014} and Tran et al.~\cite{Tran:2015}, the authors incorporated it by using optical flow. Known as two-stream network, their architecture is composed of two stream of data: (i) spatial network, which takes as input the raw RGB pixels; and (ii) temporal network, which takes as input dense optical flow displacements computed across the frames. Final predictions are computed as the average of the output scores from the two streams, showing significant improvement over other approaches. Our method differs from them by capturing not only the displacement but also velocity information provided by optical flow magnitude.

By employing the aforementioned two-stream network, Wang et al.~\cite{Wang:2015} conducted experiments showing the impact on results when changing the network architecture. In addition, they also introduced some data augmentation techniques to improve the network training. To that end, the authors used three distinct architectures (ClarifaiNet~\cite{Zeiler:2014}, GoogLeNet~\cite{Szegedy:2015} and VGG-16~\cite{Simonyan:2014b}) showing that the best results are achieved by VGG-16 deeper architecture. Afterwards, the authors improved it to the Temporal Segment Networks (TSN)~\cite{Wang:2016} by studying different types of input modalities to two-stream and by employing the Inception with batch normalization network architecture~\cite{Ioffe:2015}.

Perez et al.~\cite{Perez:2017} used MPEG motion vectors~\cite{Richardson:2003} as a different input for a two-stream network to explore temporal information. Such vectors are used to perform motion estimation in video compression where pixels are grouped in macroblocks and motion vectors are then computed for each block. They show that both optical flow and MPEG motion vectors provide equivalent accuracies, but the latter allows a more efficient implementation.


To make a spatial network learn to relate which parts of the image are moving, Park et al.~\cite{Park:2016} proposed a feature amplification technique by using magnitude information of the optical flow on the spatial network. To that end, they extract features maps of the last convolutional layer of the spatial network, compute optical flow magnitudes and resize it to be the same size of the previously extracted feature maps. Finally, they perform element-wise product to amplify the activations. Our work differs from them in that we use the magnitude information right on the beginning of the network, letting it learn how the velocity information contributes on the activity recognition process.

As it can be inferred from the reviewed methods, most of them use either convolution operations over raw pixels or optical flow to model temporal information. The former do not decouple spatial and temporal information, letting appearance information prevail~\cite{Feichtenhofer:2016}, while the latter approaches rely on horizontal and vertical components of the optical flow. Despite the optical flow-based methods produce promising results, they focus only on displacement information. In view of that, aiming at capturing more information from the optical flow, our method captures not only the displacement, by using orientation, but also captures the magnitude providing information regarding the velocity of the movement.

\section{Proposed Approach}\label{approach}


In this section, we present our approach for performing activity recognition with our proposed \emph {Magnitude-Orientation Stream} (MOS). For completeness, we first present the basic concepts of the Very Deep Two-Stream~\cite{Wang:2015}, which is the network architecture we use to learn the data representation based on the magnitude and orientation. Then, we detail our method showing how to incorporate magnitude and orientation as temporal information for the network input.

\subsection{Very Deep Two-Stream}

Motivated by the successful results achieved by deep architectures (e.g., VGG-16) in object recognition task, Wang et al.~\cite{Wang:2015} improved the two-stream network by adapting it to use the VGG-16 on activity recognition, which they called Very Deep Two-Stream CNN. As mentioned on Section~\ref{related}, the two-stream network is composed by two different networks receiving distinct flows of data, spatial and temporal. The spatial stream receives as input the RGB frames while the temporal stream receives an optical flow image as input.

The spatial network is built on a single frame image and, therefore, its architecture is the same as those for object recognition on the image domain. Thus, at each iteration of the training step, 256 training videos are uniformly sampled across the classes and a single frame is randomly selected. Moreover,  to avoid overfitting, the authors employ two data augmentation techniques: (i) cropping and flipping four corners and the center of the frame; and (ii) a multi-scale cropping method than randomly sampling the cropping width and height from {256, 224, 192, 168}. Finally, they resize the cropped regions to $224\times224\times3$.

The temporal network receives images of optical flow as input. The process for computing the optical flow is explained  as follows. For each frame $F$ on time $t$, optical flow $O_{t}$ is computed considering $F_{t}$ and $F_{t+1}$. The resulting optical flow $O_{t}$ is composed by two channels: (i) ${\mathcal O^{x}_{t}}$, denoting an image containing the x (horizontal) displacement field; and (ii) ${\mathcal O^{y}_{t}}$, denoting an image containing the y (vertical) displacement  field. Moreover, to avoid storing the displacement fields as floats, the horizontal and vertical components of the flow are linearly rescaled to a [0, 255] interval as
\begin{equation}\label{eq:linearflow}
	\mathcal I^{f}_{t_{i,j}} = \left\{
	\begin{array}{rl}
		0, &\mbox{ if $\mathcal O^{f}_{t_{i,j}} < l$} \\
		255, &\mbox{  if $\mathcal O^{f}_{t_{i,j}} > h$ } \\
		255 \times \frac{(O^{f}_{t_{i,j}} - l)}{(h - l)}, &\mbox{ otherwise }
	\end{array} \right.,
\end{equation}

\noindent where $f$ represents the image channel (flow component $x$ or $y$), $h$ is the higher bound maximum optical flow value, $l$ is the lower bound minimum optical flow value and $\mathcal I^{f}$ the optical flow image. The same data augmentation techniques used in spatial network are used in the temporal stream. Finally, the input of the temporal network is composed by stacking $10$ randomly images $\mathcal I^{f}$ of optical flow fields ($224\times224\times20$)~\cite{Simonyan:2014}. 

To perform the combination of the two networks, a late fusion scheme is employed by using a weighted linear combination of their prediction scores, where the weight is set as $2$ for temporal network and $1$ for spatial network, giving, therefore, more importance to the temporal information. Figure~\ref{img:pipeline}(a) illustrates the Deep Two-Stream network.


\subsection{Magnitude-Orientation Stream}

Our Magnitude-Orientation Stream (MOS) follows the same fundamentals as the Very Deep Two-Stream. However, aiming at extracting more information from the optical flow, MOS captures the displacement information by using orientation of the optical flow and the velocity of the movement considering the optical flow magnitude. The spatial relationship contained on local neighborhoods of magnitude and orientation captures not only displacement, by using orientation, but also magnitude providing information regarding the velocity of the movement. The method is based on non-linear transformations on the optical flow components aiming to generate input images for the temporal stream. 
%
%
To incorporate such information on the temporal stream, we compute the dense optical flow as in~\cite{Wang:2015}. In this way, for each video composed by $n$ frames, we compute $n-1$ optical flows $\mathcal O$. Once the optical flow is available, we compute the magnitude and orientation information as
\begin{equation}\label{eq:mag}
	M_{i,j} = \sqrt{ (\mathcal O^{x}_{i,j})^{2} + (\mathcal O^{y}_{i,j})^{2}}
\end{equation}
and
\begin{equation}\label{eq:ori}
	\theta_{i,j} = tan^{-1} \left( \frac{\mathcal O^{y}_{i,j}}{\mathcal O^{x}_{i,j}} \right),
\end{equation}
\noindent where $M$ and $\theta$ are the magnitude and orientation information, respectively.

\begin{figure}[!t]
	\centering
	\begin{tabular}{cc}
		\includegraphics[width=0.2\textwidth]{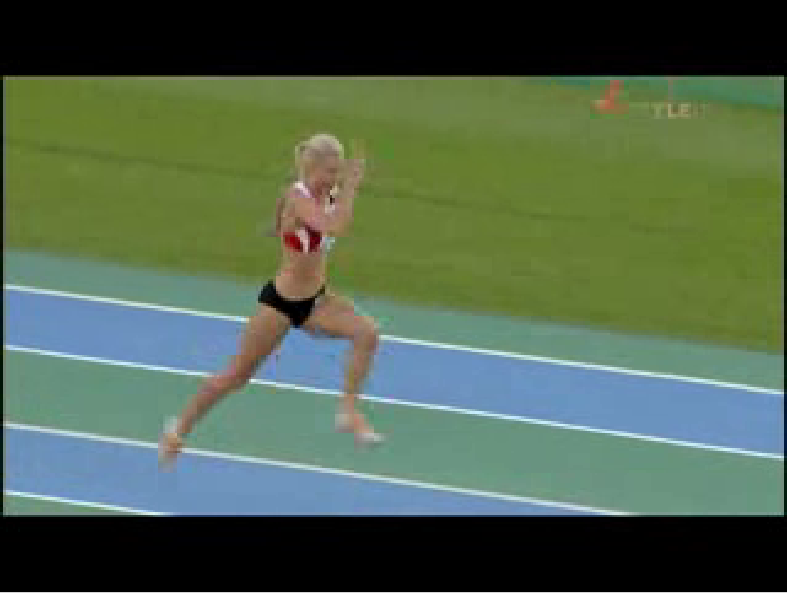} & \includegraphics[width=0.2\textwidth]{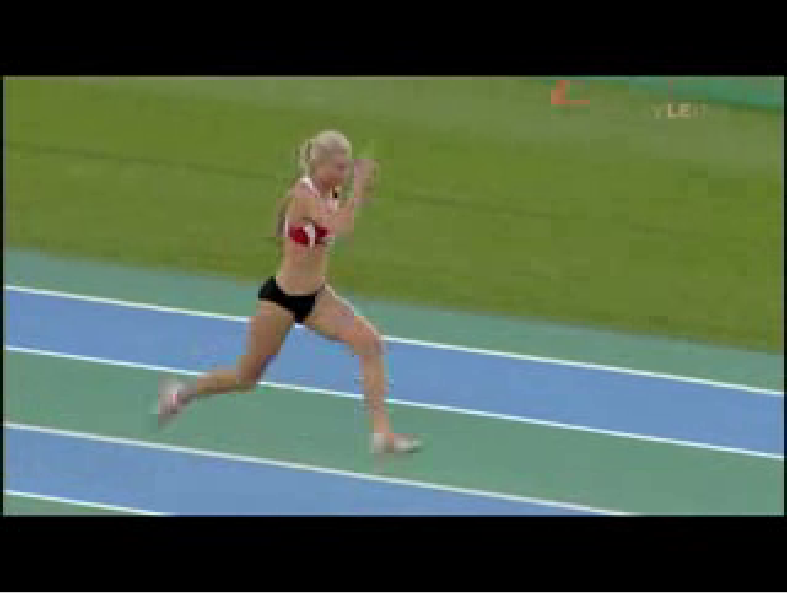} \\
		\footnotesize{(a) Frame $t$} & \footnotesize{(b) Frame $t+1$} \\
		\includegraphics[width=0.2\textwidth]{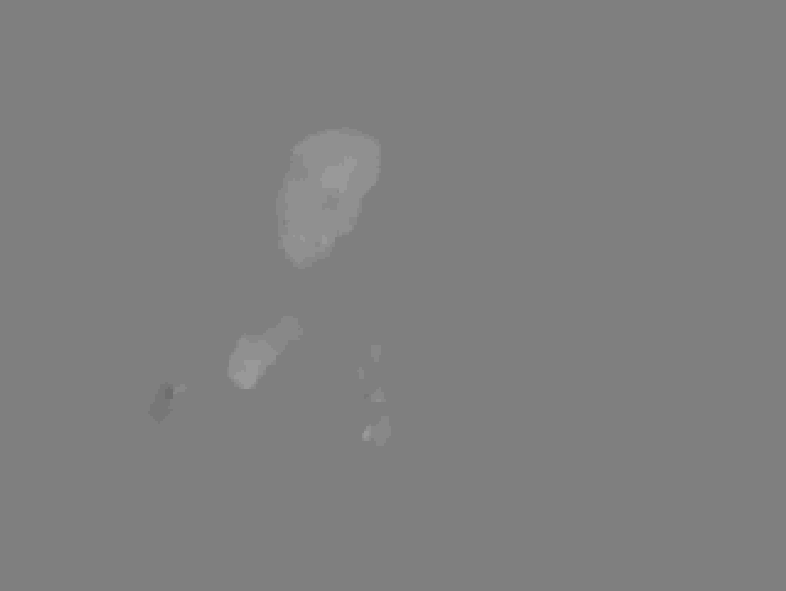} & \includegraphics[width=0.2\textwidth]{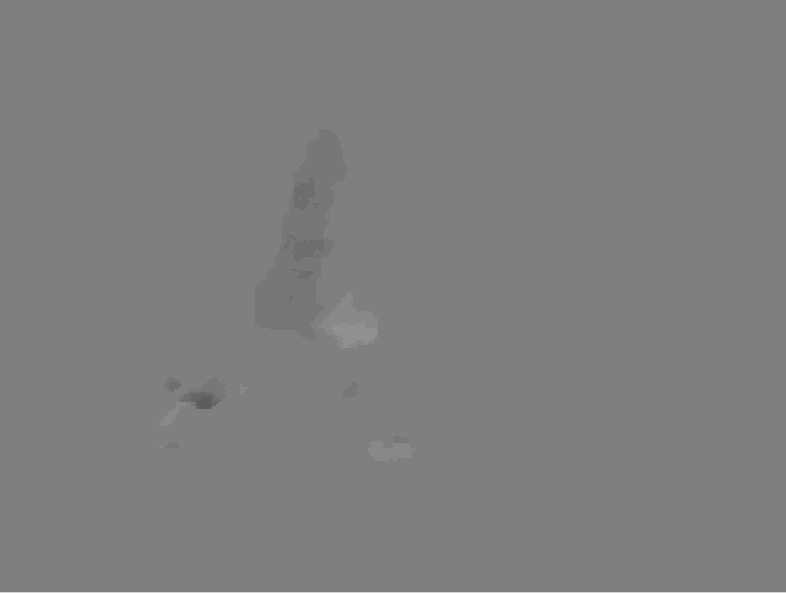} \\
		\footnotesize{(c) Horizontal displacement (flow x)} & \footnotesize{(d) Vertical displacement (flow y)} \\
		\includegraphics[width=0.2\textwidth]{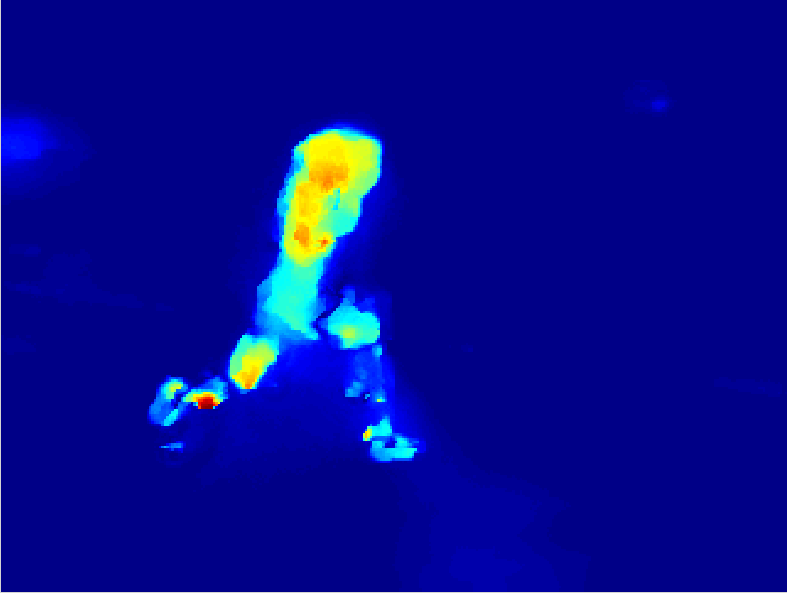} & \includegraphics[width=0.2\textwidth]{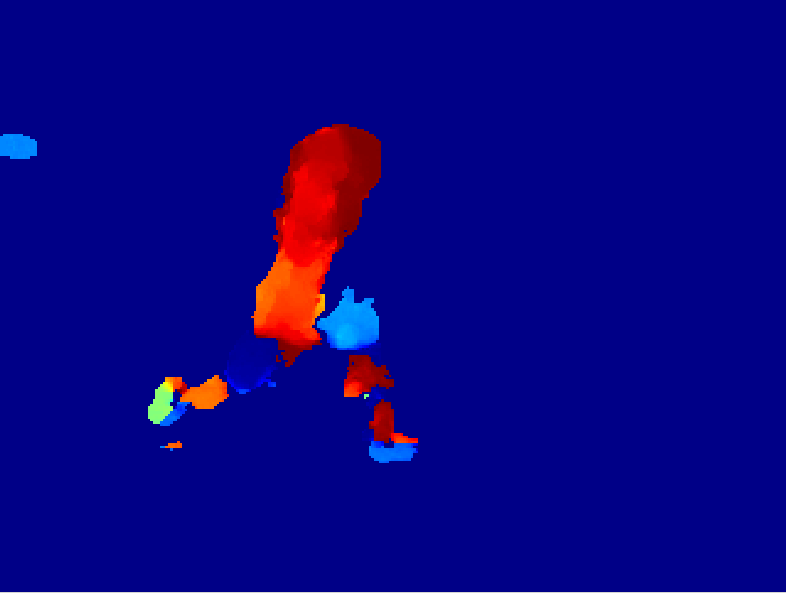} \\
		\footnotesize{(e) Magnitude ($M$)} & \footnotesize{(f) Orientation ($\theta^{'}$)} \\
	\end{tabular}
	\caption{Comparison between optical flow displacement information and magnitude and orientation extracted from tow consecutive frames ($t$ and $t+1$) of an activity sample extracted from the UCF101 dataset~\cite{Soomro:2012}.}
	\label{img:orimag}
\end{figure}

Since the values obtained in $M$ and $\theta$ are composed by real numbers, they are linearly rescaled to a [0, 255] using  Equation~\ref{eq:linearflow}. Moreover, since the orientation values are estimated for every pixel of the optical flow, it can generate noisy values of regions of the image without any movement. Therefore, we performed a filtering on $\theta$ based on the values of $M$ as 
\begin{equation}\label{eq:orifilter}
	\theta^{'}_{i,j} = \left\{
	\begin{array}{rl}
		0, &\mbox{ if $M_{i,j} < m$} \\
		\theta_{i,j}, &\mbox{ otherwise }
	\end{array} \right.,
\end{equation}
\noindent where $m$ is a magnitude threshold value. Figure~\ref{img:orimag} illustrates a comparison between the magnitude and orientation information with the optical flow x and y displacements extracted from two consecutive frames.

With the rescaled magnitude and orientation information, which can be seen as two image channels, we use the same data augmentation techniques as in~\cite{Wang:2015}. Therefore, the input is composed by 10 stacked images ($224\times224\times20$). Figure~\ref{img:pipeline}(b) illustrates the Magnitude-Orientation Stream network stages.

Finally, to incorporate spatial information to our approach, we employ a late fusion technique with the Very Deep Two-Stream network~\cite{Wang:2015}, as illustrated in Figure~\ref{img:pipeline}(c).

\section{Experimental Results}\label{experiments}

\begin{figure*}[!htb]
	\centering
	\begin{tabular}{ccc}
		\includegraphics[width=0.31\textwidth]{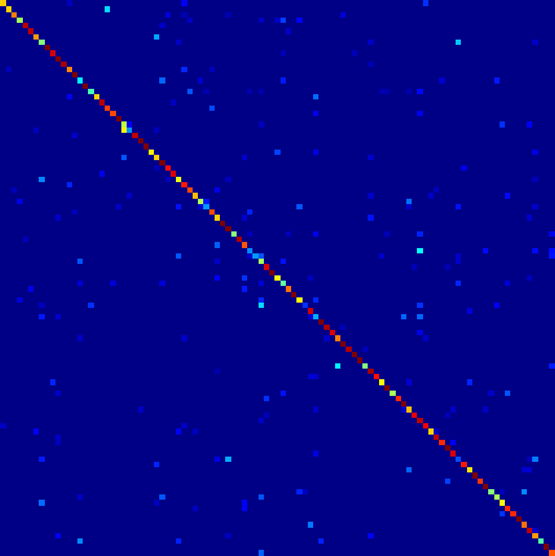} & \includegraphics[width=0.31\textwidth]{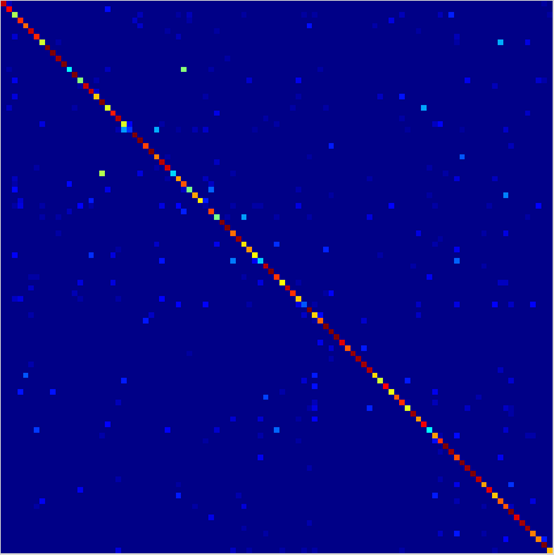} & \includegraphics[width=0.31\textwidth]{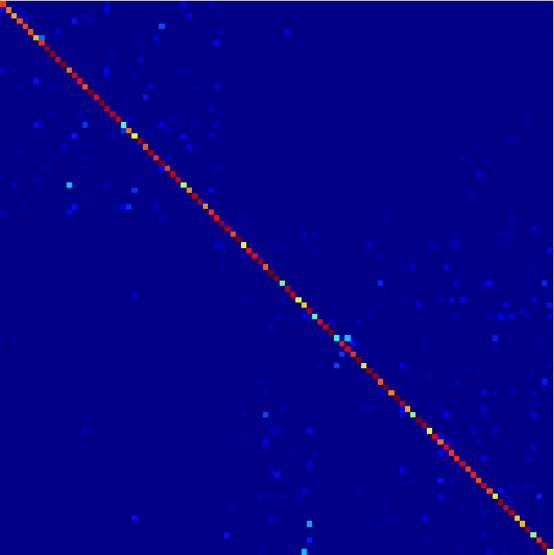}\\
		\footnotesize{(a) Very Deep Spatial Stream} & \footnotesize{(b) Very Deep Temporal Stream} & \footnotesize{(c) Magnitude-Orientation Stream (MOS)}\\
	\end{tabular}
	\caption{Confusion matrices on UCF101 split 1. False positives and false negatives were highlighted to show where each method fails.}
	\label{img:confusion_matrices}
\end{figure*}

This section describes the experimental results obtained with the proposed method for the activity recognition problem and performs comparisons to our baseline, the Very Deep Two-Stream network~\cite{Wang:2015}. To isolate only the contribution brought by our method to the activity recognition problem, the baseline was tested on the same datasets with the same split of training and testing data. The evaluations are performed considering two well-known datasets for the activity recognition problem, the UCF101~\cite{Soomro:2012} and the HMDB51~\cite{Kuehne:2011}, in which we employ the evaluation protocols and metrics proposed by their authors.

\subsection{Datasets}

The \emph{UCF101}~\cite{Soomro:2012} is an activity recognition dataset composed by videos collected from YouTube. It has a large diversity of activities and the presence of large variations in camera motion, object viewpoint, appearance, pose and scale, cluttered background, and illumination conditions. There are $13,320$ videos from $101$ activity categories grouped into $25$ groups. Each group can consist of 4-7 videos of an activity. The videos from the same group may share some common features, such as similar background or similar viewpoint. We follow the original protocol using three train-test splits. The performance is evaluated by computing the average recognition across all classes over the three splits as in~\cite{Wang:2015}.

The \emph{HMDB51}~\cite{Kuehne:2011} is a realistic and challenging activity dataset composed of video clips from movies, the Prelinger archive, Internet, Youtube and Google videos, and comprised of $51$ activity categories. It consists of $6,766$ activity samples with a resolution of $240$ pixels in height with preserved aspect ratio. We follow the original protocol using three train-test splits. The performance is evaluated by computing the average accuracy across all classes over the three splits.

\subsection{Implementation Details}

\subsubsection{Network Architecture}

We employed the VGG-16 deep convolutional network~\cite{Simonyan:2014b}. It is a convolutional architecture composed by $13$ convolutional and three fully-connected layers with smaller convolutional size, stride and pooling window ($3 \times 3$, $1 \times 1$ and $2 \times 2$, respectively). We opted for using such architecture since it demonstrated the best performance in~\cite{Wang:2015}. In addition, we used a model compatible with a modified version of the Caffe framework~\cite{Jia:2014} made available\footnote{\url{https://github.com/yjxiong/caffe}}.

\subsubsection{Pre-training}

As stated by~\cite{Wang:2015}, the UCF101 dataset training split is very small to train a deep convolutional network (the same applies to HMDB51, since it has less videos). In view of that, a possible solution used by several works~\cite{Simonyan:2014, Wang:2015, Feichtenhofer:2016, Wang:2016} is to use
ImageNet models as the initialization for network training. In this way, here we also employed the ImageNet model as pre-training

\subsubsection{Training} 

Following the implementation details used by our baseline~\cite{Wang:2015}, we set the learning rate initially to $0.005$ and then decreases it at every $5,000$ iterations dividing it by $10$. The maximum iteration was set as $15,000$.
We kept the same schedule for all training sets. 

Similarly to~\cite{Simonyan:2014, Wang:2015}, the network weights are learned using the mini-batch stochastic gradient descent with a momentum set to $0.9$ and weight decay of $0.0005$. We also set high dropout ratio for the fully connected layers ($0.9$ and $0.8$).

Krizhevsky et al.~\cite{Krizhevsky:2012} demonstrated that data augmentation techniques can be very effective to avoid overfitting. In view of that, we cropped and flipped four corners and the center of the frame. In addition, we applied a multi-scale cropping method and randomly sampled the cropping width and height from $\{256, 224, 192, 168\}$ (finally, we resize the cropped regions to $224 \times 224$). It is important to state that our baseline~\cite{Wang:2015} used the same data augmentation procedure.

\subsubsection{Test} 

To perform a fair comparison, we applied the same test scheme used by our baseline~\cite{Wang:2015}, described as follows. First, we sample $25$ magnitude/orientation flow images for the testing. Then, from each of these, we obtain 10 convolutional network inputs (by cropping and flipping four corners and the center). Finally, the prediction score for the input video is obtained by averaging the sampled images scores and their crops. The same testing scheme was used by the original two-stream convolutional network~\cite{Simonyan:2014}. For the fusion of MOS and other streams, we use a non-weighted linear combination of their prediction scores.

\subsubsection{Optical Flow Extraction}

As mentioned on Section~\ref{approach}, the magnitude/orientation images are computed from the optical flow information. To that end, we extract the optical flow information using the TVL1 algorithm~\cite{Zach:2007}, implemented in OpenCV with CUDA. For the sake of comparison, our baseline~\cite{Wang:2015} used the same optical flow algorithm. To obtain the magnitude and orientation images information we empirically set the parameters $h = 15$ and $l = -15$ to compute $M$; and $h = 180$, $l = -180$ and $m = 128$ to compute $\theta^{'}$.

\subsection{Evaluation}

\begin{table*}[th!]
	\centering
	\begin{small}
		\caption{Activity recognition accuracy (\%) results of Magnitude-Orientation Stream and the baseline on UCF101~\cite{Soomro:2012} activity dataset. Results for the baseline were obtained running the code provided by the authors~\cite{Wang:2015}. Note that our results were achieved with only our single Magnitude-Orientation Stream (temporal information) while the results of~\cite{Wang:2015} consider two streams (spatial and temporal information).}
		\begin{tabular}{clp{1.5cm}p{1.5cm}p{1.5cm}p{1.5cm}}
			\toprule
			& & \multicolumn{1}{c}{\textbf{Split 1}} & \multicolumn{1}{c}{\textbf{Split 2}} & \multicolumn{1}{c}{\textbf{Split 3}} & \multicolumn{1}{c}{\textbf{Average}}\vspace{0.075cm}\\
			& \textbf{Approach} & \multicolumn{1}{c}{\textbf{Acc. (\%)}} & \multicolumn{1}{c}{\textbf{Acc. (\%)}} & \multicolumn{1}{c}{\textbf{Acc. (\%)}} & \multicolumn{1}{c}{\textbf{Acc. (\%)}}\\
			\toprule
			& Very Deep Spatial Stream~\cite{Wang:2015} & \multicolumn{1}{c}{79.8} & \multicolumn{1}{c}{77.3} & \multicolumn{1}{c}{77.8} & \multicolumn{1}{c}{78.4} \\
			\multirow{1}{*}{\textbf{Baseline}} & Very Deep Temporal Stream~\cite{Wang:2015} & \multicolumn{1}{c}{85.7} & \multicolumn{1}{c}{88.2} & \multicolumn{1}{c}{87.4} & \multicolumn{1}{c}{87.0} \\ 
			& Very Deep Two-Stream~\cite{Wang:2015} & \multicolumn{1}{c}{90.9} & \multicolumn{1}{c}{91.6} & \multicolumn{1}{c}{91.6} & \multicolumn{1}{c}{\textbf{91.4}} \\
			\midrule
			
			& Magnitude-Orientation Stream (MOS) & \multicolumn{1}{c}{90.8} & \multicolumn{1}{c}{{89.3}} & \multicolumn{1}{c}{91.5} & \multicolumn{1}{c}{{90.5}} \\
			\multirow{1}{*}{\textbf{Our}} & Magnitude-Orientation Stream (MOS) + Very Deep Spatial Stream & \multicolumn{1}{c}{93.1} & \multicolumn{1}{c}{{91.9}} & \multicolumn{1}{c}{92.6} & \multicolumn{1}{c}{{\textbf{92.5}}} \\
			\multirow{1}{*}{\textbf{results}} & Magnitude-Orientation Stream (MOS) + Very Deep Temporal Stream & \multicolumn{1}{c}{91.4} & \multicolumn{1}{c}{{92.2}} & \multicolumn{1}{c}{93.6} & \multicolumn{1}{c}{{\textbf{92.4}}} \\
			& Magnitude-Orientation Stream (MOS) + Very Deep Two-Stream& \multicolumn{1}{c}{93.7} & \multicolumn{1}{c}{{93.1}} & \multicolumn{1}{c}{94.8} & \multicolumn{1}{c}{{\textbf{93.8}}} \\
			\bottomrule
		\end{tabular}
		\label{tab:baseline-comparison}
	\end{small}
\end{table*}


We report the activity recognition performance of our Magnitude-Orientation Stream in contrast with the baseline in Table~\ref{tab:baseline-comparison} showing a comparison of our method to the three different streams of our baselines (Very Deep Spatial Stream, Very Deep Temporal Stream and Very Deep Two-Stream). A considerable improvement was obtained with Magnitude-Orientation Stream when compared to the baseline single streams, reaching $90.8\%$ of accuracy on split 1 of the UCF101 dataset. There is an improvement of $5.1$ percentage points (p.p.) when compared to the Very Deep Temporal Stream~\cite{Wang:2015} and $11.0$ p.p. when compared to the Very Deep Spatial Stream~\cite{Wang:2015}. Furthermore, it is worth noting that our best result using Magnitude-Orientation Stream on split 1 is close to the best one reported (Very Deep Two-Stream) which is obtained by using a combination of two different streams (spatial and temporal informations), while we only used our single Magnitude-Orientation Stream (temporal information). The same observations can be considered when analyzing the results of our temporal stream on splits 2 and 3. Therefore, such results can be considered remarkably good and confirm the advantages introduced by our approach.


Figure~\ref{img:confusion_matrices} shows the confusion matrices of Very Deep Spatial Stream, Very Deep Temporal Stream and our Magnitude-Orientation Stream for the UCF101 split 1 (we highlighted the false positives and false negatives to make it more visible on where each method fails). We can observe that our approach fails on classes that are more semantically closer to each other\footnote{Since the actions on the confusion matrices are sorted according to its labels (e.g., ApplyEyeMakeup, ApplyLipstick, or BaseballPitch, Basketball, BasketballDunk), near regions in the confusion matrix denote  semantically closer activities.}, whereas the Very Deep Spatial Stream and the Very Deep Temporal Stream fails in a random manner. In addition, the three methods produce false positives and false negatives different from each other, indicating the possibility of fusion.

\begin{table}[!htb]
	\centering
	\begin{small}
		\caption{Activity recognition accuracy (\%) results of handcrafted methods, neural networks (NN) methods and our Magnitude-Orientation Stream on UCF101 activity dataset~\cite{Soomro:2012}. Results for features + BoW were obtained from~\cite{Shi:2015} and features + FV were obtained from~\cite{Shi:2014:Dissertation}.}
		\begin{tabular}{clc}
			\toprule
			& & \multicolumn{1}{c}{\textbf{UCF101}} \\
			& \textbf{Approach} & \multicolumn{1}{c}{\textbf{Acc. (\%)}} \\
			\toprule
			& HOF + BoW~\cite{Laptev:2008} & \multicolumn{1}{c}{61.8}  \\ 
			& HOG-HOF + BoW~\cite{Laptev:2008} & \multicolumn{1}{c}{71.8}  \\ 
			& MBH + BoW~\cite{Dalal:2006} & \multicolumn{1}{c}{77.1}  \\ 
			& GBH + BoW~\cite{Shi:2015} & \multicolumn{1}{c}{68.5}  \\ 
			& HOG3D + BoW~\cite{Klaser:2008} & \multicolumn{1}{c}{61.4}  \\ 
			
			& HOF + FV~\cite{Laptev:2008} & \multicolumn{1}{c}{65.9} \\ 
			\multirow{1}{*}{\textbf{Handcrafted}} & HOG-HOF + FV~\cite{Laptev:2008} & \multicolumn{1}{c}{75.4}  \\ 
			\multirow{1}{*}{\textbf{Methods}} & MBH + FV~\cite{Dalal:2006} & \multicolumn{1}{c}{81.0}  \\  
			& GBH + FV~\cite{Shi:2015} & \multicolumn{1}{c}{74.2} \\  
			& HOG3D + FV~\cite{Klaser:2008} & \multicolumn{1}{c}{64.7}  \\ 
			
			& IDT~\cite{Wang:2013} & \multicolumn{1}{c}{85.9}  \\ 
			& IDT + higher FV~\cite{Peng:2016} & \multicolumn{1}{c}{87.9} \\
			& IDT + MVSV~\cite{Cai:2014} & \multicolumn{1}{c}{83.5} \\
			\midrule
			
			& Deep Networks~\cite{Karpathy:2014} & \multicolumn{1}{c}{65.4} \\
			& Composite LSTM~\cite{Srivastava:2015} & \multicolumn{1}{c}{75.8} \\
			& C3D~\cite{Tran:2015} & \multicolumn{1}{c}{85.2}  \\
			\multirow{1}{*}{\textbf{NN}} & Factorized CNN~\cite{Sun:2015} & \multicolumn{1}{c}{88.1}  \\
			\multirow{1}{*}{\textbf{Methods}} & Two-Stream~\cite{Simonyan:2014} & \multicolumn{1}{c}{88.0}\\ 
			& Two-Stream F~\cite{Feichtenhofer:2016} & \multicolumn{1}{c}{92.5}  \\
			& KVMF~\cite{Zhu:2016} & \multicolumn{1}{c}{93.1} \\
			& TSN~\cite{Wang:2016} & \multicolumn{1}{c}{\textbf{94.2}} \\
			\midrule
			&\multirow{1}{*}{MOS} & \multicolumn{1}{c}{\multirow{1}{*}{{90.5}}} \\
			\multirow{1}{*}{\textbf{Our}} & \multirow{1}{*}{MOS + Very Deep Spat. Stream} & \multicolumn{1}{c}{\multirow{1}{*}{{92.5}}} \\
			\multirow{1}{*}{\textbf{results}} & \multirow{1}{*}{MOS + Very Deep Temp. Stream} & \multicolumn{1}{c}{\multirow{1}{*}{{92.4}}}  \\
			& \multirow{1}{*}{MOS + Very Deep Two-Stream} & \multicolumn{1}{c}{\multirow{1}{*}{{\textbf{93.8}}}} \\
			\bottomrule
		\end{tabular}
		\label{tab:ucf101-comparison}
	\end{small}
\end{table}

To exploit a possible complementarity of the three approaches  (very deep spatial stream, very deep temporal stream and our magnitude-orientation stream),  we combined the different streams by employing a late fusion technique using a weighted linear combination of their prediction scores. According to the results showed in Table~\ref{tab:baseline-comparison}, any type of combination performed with our Magnitude-Orientation Stream provides better results than Very Deep Two-Stream, with the best result achieving an improvement of 2.4 p.p. over Very Deep Two-Stream.

Table~\ref{tab:ucf101-comparison} presents results on UCF101 dataset for many works. The first part of the table shows results of methods that extract temporal information using handcrafted features. We compare our MOS approach with the results of local feature-based methods, such as Bag-of-Words (BoW) + features, Fisher vector (FV) + features, and Improved Dense Trajectories (IDT). The best result by such type of methods was achieved with IDT + higher FV~\cite{Peng:2016}, reaching $87.9\%$. Our best result using the proposed approach combined with Very Deep Two-Stream outperforms that by 5.9 percentage points.

\begin{table}[!htb]
	\centering
	\begin{small}
		\caption{Activity recognition accuracy (\%) results of handcrafted methods, neural networks (NN) methods and our Magnitude-Orientation Stream on HMDB51 activity dataset~\cite{Kuehne:2011}. Results for features + BoW were obtained from~\cite{Shi:2015} and features + FV were obtained from~\cite{Shi:2014:Dissertation}.}
		\begin{tabular}{clc}
			\toprule
			& & \multicolumn{1}{c}{\textbf{HMDB51}} \\
			& \textbf{Approach} & \multicolumn{1}{c}{\textbf{Acc. (\%)}} \\
			\toprule
			& HOF + BoW~\cite{Laptev:2008} & \multicolumn{1}{c}{35.5} \\ 
			& HOG-HOF + BoW~\cite{Laptev:2008}  & \multicolumn{1}{c}{43.6} \\ 
			& MBH + BoW~\cite{Dalal:2006}  & \multicolumn{1}{c}{51.5} \\ 
			& GBH + BoW~\cite{Shi:2015}  & \multicolumn{1}{c}{38.8} \\ 
			& HOG3D + BoW~\cite{Klaser:2008}  & \multicolumn{1}{c}{36.2} \\ 
			& OFCM + BoW~\cite{Caetano:2016}  & \multicolumn{1}{c}{56.9} \\ 
			
			\multirow{1}{*}{\textbf{Handcrafted}} & HOF + FV~\cite{Laptev:2008}  & \multicolumn{1}{c}{39.9} \\ 
			\multirow{1}{*}{\textbf{Methods}} & HOG-HOF + FV~\cite{Laptev:2008}  & \multicolumn{1}{c}{45.6} \\ 
			& MBH + FV~\cite{Dalal:2006} & \multicolumn{1}{c}{54.7} \\  
			& GBH + FV~\cite{Shi:2015}  & \multicolumn{1}{c}{44.7} \\  
			& HOG3D + FV~\cite{Klaser:2008} & \multicolumn{1}{c}{38.2} \\ 
			
			& IDT~\cite{Wang:2013} & \multicolumn{1}{c}{57.2} \\ 
			& IDT + higher FV~\cite{Peng:2016}  & \multicolumn{1}{c}{61.1} \\
			& IDT + MVSV~\cite{Cai:2014} & \multicolumn{1}{c}{55.9} \\
			\midrule
			
			& Composite LSTM~\cite{Srivastava:2015}  & \multicolumn{1}{c}{44.0} \\
			& Factorized CNN~\cite{Sun:2015} & \multicolumn{1}{c}{59.1} \\
			\multirow{1}{*}{\textbf{NN}} & Two-Stream~\cite{Simonyan:2014}  & \multicolumn{1}{c}{59.4} \\ 
			\multirow{1}{*}{\textbf{Methods}} & Two-Stream F~\cite{Feichtenhofer:2016}  & \multicolumn{1}{c}{65.4} \\
			& KVMF~\cite{Zhu:2016} & \multicolumn{1}{c}{63.3} \\
			& TSN~\cite{Wang:2016} & \multicolumn{1}{c}{\textbf{69.4}} \\
			\midrule
			\multirow{1}{*}{\textbf{Our}} &\multirow{2}{*}{MOS} & \multicolumn{1}{c}{\multirow{2}{*}{\textbf{66.2}}} \\
			\multirow{1}{*}{\textbf{results}} & & \\
			\bottomrule
		\end{tabular}
		\label{tab:hmdb51-comparison}
	\end{small}
\end{table}

The second part of Table~\ref{tab:ucf101-comparison} shows results achieved with neural networks (NN) approaches. 
According to the results, just using our Magnitude-Orientation Stream (MOS), we outperform many methods (\cite{Karpathy:2014, Srivastava:2015, Tran:2015, Sun:2015, Simonyan:2014}). In comparison with C3D~\cite{Tran:2015}, we outperform them by 5.3 p.p. using our temporal stream and 8.6 p.p. when combining it with Very Deep Two-Stream. This indicates that our magnitude orientation approach learns temporal information better than the approaches that perform 3D convolution operations directly. Moreover, such finding is very important since 3D convolutional operations are more computationally expensive than the 2D convolutional operations used in our approach. 
It is worth mentioning that we also improved the results achieved by the original two-stream~\cite{Simonyan:2014} by 2.5 p.p. using our temporal stream and by 5.8 p.p. combining it with Very Deep Two-Stream. Finally, we can observe that our best result only did not outperform Wang et al. TSN method~\cite{Wang:2016}, however it presents results very close to it.

Finally, Table~\ref{tab:hmdb51-comparison} presents the results achieved on HMDB51 dataset with works of the literature that tried to handle temporal information by handcrafted features (first part) as well as recent works that employ neural networks (second part). Again, the best handcrafted method result is achieved by IDT + higher FV~\cite{Peng:2016} reaching $61.1\%$. Our MOS approach outperformed it by $5.1$ p.p. When compared with neural network methods, we were able to outperform many methods, e.g.,~\cite{Karpathy:2014, Srivastava:2015, Tran:2015, Sun:2015, Simonyan:2014, Zhu:2016}, using the proposed Magnitude-Orientation Stream (MOS). Furthermore, we were able to outperform the original two-stream~\cite{Simonyan:2014} by $6.6$ p.p. just using our temporal stream. Once more, our approach only did not outperform TSN~\cite{Wang:2016}. Here. we believe this is because HMDB51 is a smaller dataset lacking training data\footnote{The same statement was made by Simonyan and Zisserman~\cite{Simonyan:2014}, which evaluated different options for increasing the effective training set size of HMDB51.}.

\section{Conclusions and Future Works}\label{conclusions}

In this work, we proposed a novel temporal stream for two-stream convolutional networks, named Magnitude-Orientation Stream (MOS). The method is based on simple non-linear transformations on the optical flow components generating input images composed of magnitude and orientation information. The spatial relationship contained on local neighborhoods of magnitude and orientation captures not only displacement, by using orientation, but also magnitude providing information regarding the velocity of the movement. We demonstrated that MOS outperforms all classic approaches based on local handcrafted features of the literature. Furthermore, simply by using only our temporal stream, we outperform original CNN two-stream approaches based on temporal and spatial information as well as other recent works that employ neural networks, suggesting its suitability to learn temporal information. Another interesting finding is that the combination of our temporal stream with the Very Deep Two-Stream method improves the activity recognition.

Directions to future works include the evaluation of the magnitude-orientation information with other distinct architectures, such as the Inception with batch normalization. Moreover, we intend to evaluate its behavior on other video-related problems, such as action detection or scene description.

\section*{Acknowledgments}

The authors would like to thank the Brazilian National Research Council -- CNPq, the Minas Gerais Research Foundation -- FAPEMIG (Grants APQ-00567-14 and PPM-00540-17) and the Coordination for the Improvement of Higher Education Personnel -- CAPES (DeepEyes Project). The authors gratefully acknowledge the support of NVIDIA Corporation with the donation of the GeForce Titan X GPU used for this research.

\balance
\bibliographystyle{IEEEtran}
\bibliography{main}

\end{document}